\pdfoutput=1

\documentclass[11pt]{article}

\usepackage[preprint]{acl}

\usepackage{times}
\usepackage{latexsym}
\usepackage{amsmath}

\usepackage[T1]{fontenc}

\usepackage[utf8]{inputenc}
\usepackage{booktabs}
\usepackage{microtype}

\usepackage{inconsolata}

\usepackage{graphicx}
\usepackage{expex}
\usepackage{linguex}
\title{Meaning Beyond Truth Conditions: Evaluating Discourse Level Understanding via Anaphora Accessibility}

\author{Xiaomeng Zhu${}^*$
\hspace{0.4in} Zhenghao Zhou${}^*$ \hspace{0.4in} Simon Charlow \hspace{0.4in} Robert Frank \\ Department of Linguistics \\ Yale University\\ \texttt{\{miranda.zhu, herbert.zhou, simon.charlow, robert.frank\}@yale.edu}}

\begin{document}
\maketitle

\def\thefootnote{*}\footnotetext{Equal contribution.}
\begin{abstract}
We present a hierarchy of natural language understanding abilities and argue for the importance of moving beyond assessments of understanding at the lexical and sentence levels to the discourse level.  We propose the task of \textit{anaphora accessibility} as a diagnostic for assessing discourse understanding, and to this end, present an evaluation dataset inspired by theoretical research in dynamic semantics. We evaluate human and LLM performance on our dataset and find that LLMs and humans align on some tasks and diverge on others. Such divergence can be explained by LLMs' reliance on specific lexical items during language comprehension, in contrast to human sensitivity to structural abstractions.
\end{abstract}

\section{Introduction} \label{sec:intro}

The success of modern large language models (LLMs) depends on their capacity for natural language understanding (NLU), i.e., the ability to extract the semantic information contained in a text. Systematic assessment of NLU abilities has been carried out using a diverse set of  evaluation tasks, but few of them target whether LLMs accurately represent and update states of natural language discourse. Successful interpretation of discourse requires the ability to use pronominal expressions to refer to entities that have been introduced in a text.

\begin{figure}[t]
    \centering
    \includegraphics[width=1\linewidth]{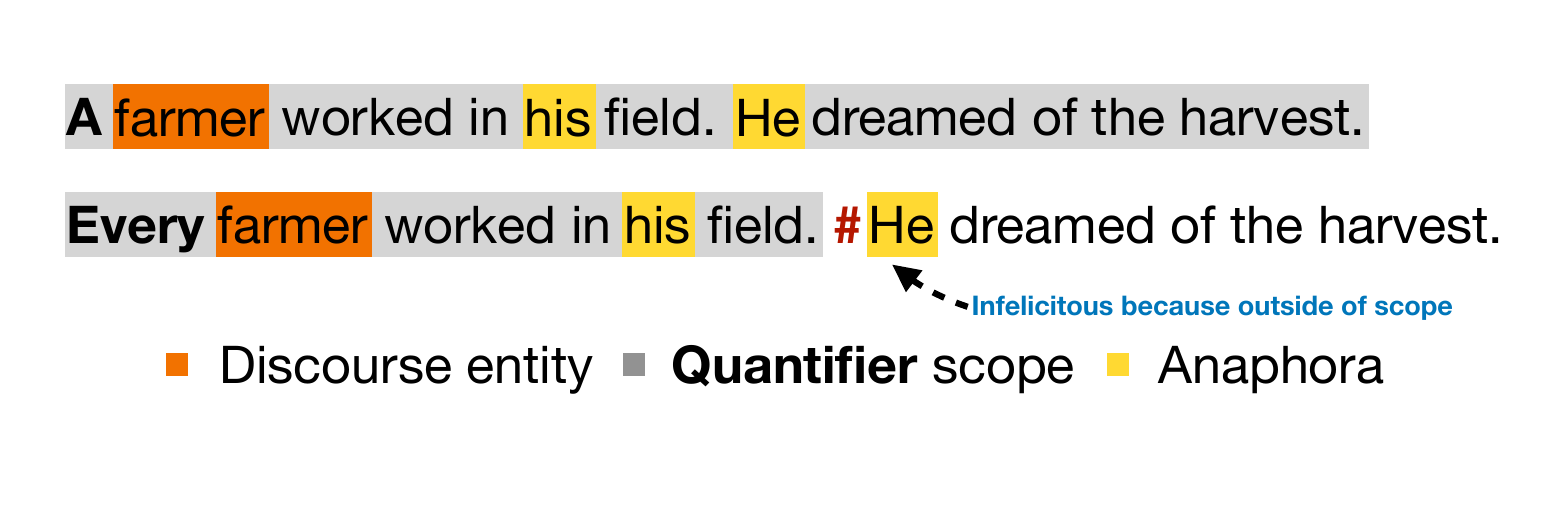}
    \caption{Quantifier scope and its impact on anaphora. 
    }
    \label{fig:scopefig}
\end{figure}

The felicity of \textbf{pronominal anaphora}, i.e., using pronouns to refer back to discourse referents introduced earlier, is influenced by the semantic scope of the antecedent:

\pex[aboveexskip=4pt, belowexskip=4pt]<demo> 
\{A, \#Every\} farmer worked in his field. He dreamed of the harvest.
\xe
Example (\getref{demo}) shows that an entity introduced by an existential quantifier is \textbf{accessible} in the same sentence, as well as in subsequent sentences. In contrast, entities introduced by universal quantifiers are only accessible to pronouns in the same sentence; anaphora is infelicitous otherwise. This is illustrated in Figure \ref{fig:scopefig}: the discourse referent is \textbf{subordinated} to the universal quantifier --- that is, inaccessible outside its scope, which extends to the end of the first sentence in the sequence. This makes subsequent reference to \textit{he} in the second sentence infelicitous.

The process of introducing discourse referents is formalized in `dynamic' variants of formal semantics (e.g., \citealp{heim1983projection, groenendijk1991dynamic, kamp2010discourse}). In dynamic semantics, utterances precipitate changes in the discourse state, for example by introducing discourse referents. This gives rise to notions of discourse or textual scope which differentiate (e.g.) existential and universal quantifiers, in line with Figure~\ref{fig:scopefig}.

Here, we focus on one aspect of discourse-level semantic knowledge, namely the fine-grained interactions between semantic scope and referent accessibility. We investigate whether LLMs demonstrate knowledge of the semantic scope properties of various quantifiers and logical connectives, and whether this knowledge is used to generate and update representations of discourse states in human-like ways. 

\begin{figure*}[h!]
    \centering
    \includegraphics[width=1\linewidth]{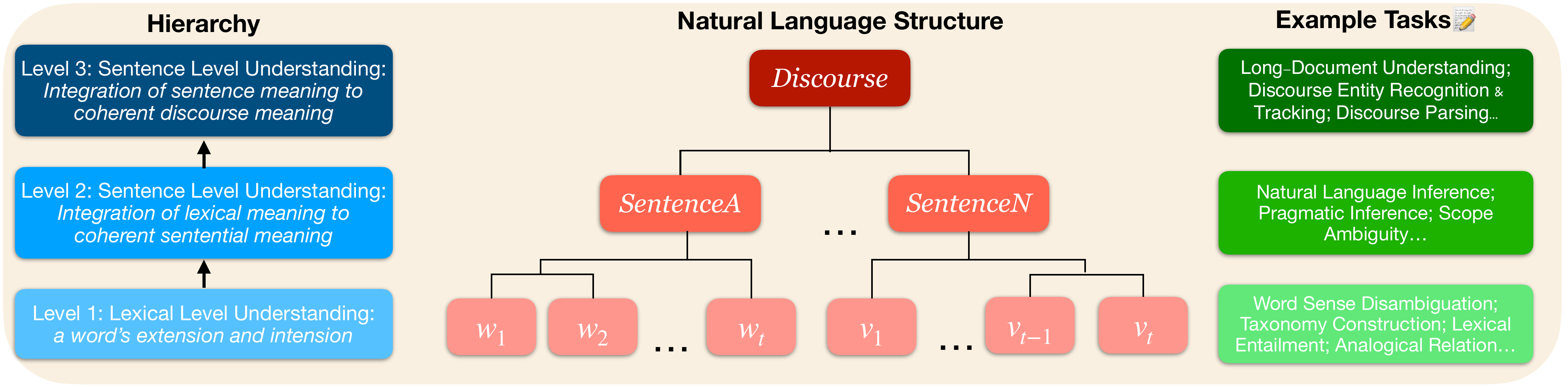}
    \caption{Proposed hierarchy of levels of semantic understanding abilities.}
    \label{fig:hierarachy}
\end{figure*}

\paragraph{Contribution} We make the following contributions:
\begin{itemize}
\setlength\itemsep{0.1em}
    \item In Section~\ref{sec:hierarchy}, we propose a hierarchy of levels of semantic understanding abilities, which can serve as a guideline for characterizing the kinds of semantic knowledge that LLMs have.
    \item In Section~\ref{sec:method_all}, we propose an evaluation dataset covering discourse anaphora across a variety of linguistic constructions, all of which require sensitivity to the way in which the form of language determines the ways discourse states are implicitly updated in natural discourse.  
    \item In Sections~\ref{sec:exp1} through \ref{sec:exp3}, we evaluate both LLMs and humans with our dataset, and uncover intriguing patterns where human and model behavior align and differ.
\end{itemize}

\section{Levels of Semantic Understanding} \label{sec:hierarchy}

Figure~\ref{fig:hierarachy} illustrates three different levels of natural language understanding: (i) lexical level, (ii) sentential level, and (iii) discourse level. Semantic competence, we propose, requires knowledge of all of these. We discuss each one in detail and review existing work that has tried to evaluate LLM capacities at that level. 

\subsection{Lexical Level}
We define lexical level understanding as \textbf{knowing the meaning of individual lexical items}. This requires knowledge of a word's extension (the objects in the world that a word picks out) and its intension (the objects it would pick out if the world were different).  
Such knowledge allows a competent speaker to make judgments of synonymy, antonymy, entailment and the like. In LLMs, lexical knowledge corresponds to vector representations of individual tokens.

\citet{moskvoretskii-etal-2024-taxollama} summarize a range of Natural Language Understanding (NLU) tasks that assess lexical level understanding: Word Sense Disambiguation, Hypernym Discovery, Taxonomy Construction, Lexical Entailment, etc.
Another test of lexical semantic understanding derives from the analogical reasoning tests explored by \citet{mikolov-etal-2013-linguistic}, where word meaning is needed to complete analogies such as \textit{man}:\textit{king} as \textit{woman}:\textit{X}. 
All of these tasks rely on  knowledge of word meaning that is independent of the effects on meaning that derives from the composition of words in phrases and sentences.

\subsection{Sentence Level}
On top of the building blocks provided by lexical understanding, sentence understanding is \textbf{the ability to integrate lexical meanings in phrases and to form coherent semantic representations for sentences}. 
Traditionally, sentence-level meaning is identified with truth conditions and encoded using a logical formalism with rigorously defined semantics (e.g., \citealp{HeimKratzer:1998}).

A model's capacity to encode the truth conditions of single sentences is implicated in important NLU tasks such as Natural Language Inference (NLI), which requires LLMs to form accurate meaning representations for two sentences and classify their logical relations as entailment, contradiction, or neutral \citep{williams-etal-2018-broad}. Similar evaluation tasks have been created for pragmatic inferences, targeting implicature and presupposition \citep{jeretic-etal-2020-natural}. These works investigate meaning representations of pairs of minimally different sentences, either with respect to logical relations or pragmatic relations, without the need to connect the two sentences in sequential order or track changes at the discourse level. Another type of work at the sentence level involves ambiguities, such as scope ambiguity (e.g., \citealp{kamath-etal-2024-scope}): a single sentence with multiple quantifiers might allow different interpretations given specific scopal arrangements between the quantifiers.

\subsection{Discourse Level}
We define discourse level understanding as \textbf{the ability to integrate the meaning of consecutive sentences into a unified discourse representation}. Discourse-level meaning requires moving beyond formalisms that express meaning as a static representation of truth conditions to dynamic formalisms in which meaning accrues via update to a contextual representation or state.

One type of task that probes discourse level understanding is discourse parsing (e.g., \citealt{maekawa-etal-2024-obtain}), which evaluates the ability of a model to determine the relationships between sentences, such as \textit{elaboration}, \textit{attribution}, etc.  While informative, this task requires the adoption of specific assumptions about the structure and categories that determine discourse relations. 

An alternative, more theory-neutral evaluation considers the accumulation of information through a discourse. \citet{li-etal-2021-implicit} examine the tracking of the state of individuals and situations across a text. They probed the internal representations of encoder-decoder transformers and found localizable, interpretable structures, supporting the claim that pretrained language models implicitly simulate entity tracking processes dynamically. \citet{kim-schuster-2023-entity} extended the paradigm in \citet{li-etal-2021-implicit} by removing the potential shortcuts that models can use in inferring the states of discourse entities. 
This line of work uses natural language to explicitly describe the initial state of a situation as well as each subsequent change in the state (e.g. \textit{Box 1 contains the book. Box 2 contains the apple.... Move the book into Box 2...}), thereby functionally similar to the core idea of dynamic semantics. However, because of the simplicity of the language involved, this task did not probe sensitivity to the specific lexical items and syntactic structures that impact the evolution of discourse state, the focus of the current work.

Another line of evaluation targets how processing each sentence in a discourse impacts the entities that can be discussed, the task of   discourse entity recognition \citep{schuster-linzen-2022-sentence, zhu-frank-2024-lieder}. \citeauthor{schuster-linzen-2022-sentence} examine sensitivity to the scope of negation at the discourse level: an indefinite interpreted within the scope of negation should not introduce an entity that can be referred to. They found that while LLMs indeed exhibit such sensitivity, their performance is not systematic. \citet{zhu-frank-2024-lieder} extended their paradigm by increasing the types of test items, which allows for the evaluation of the semantic properties that govern discourse entity introduction and reference. 
However, both \citet{schuster-linzen-2022-sentence} and \citet{zhu-frank-2024-lieder} only evaluated LLMs on sentences of a rather simple structure, such as \textit{John owns a dog but Mark does not own a dog}, which only considers negation as the scope that interacts with discourse entities. This gap in the literature calls for a more comprehensive evaluation of \textbf{other scopes} (such as existentials, universals, conditionals, and disjunctions) that interact with discourse entities, as in the present study.


\section{Evaluating Discourse-level Meaning Representation: Case Study on Anaphora (In)accessiblity}\label{sec:method_all}
As discussed in the previous section, existing work on the evaluation of LLMs' discourse level semantic understanding leaves unexplored the implications of the fine details of semantic composition and scope on the representation of discourse context.  
As we elaborate below, 
the scopal properties of quantifiers and logical connectives that are determined by sentence level semantic interpretation play a significant role in discourse level interpretation: depending on the semantic operator, they may license discourse entities only within their scope. 
We exploit such patterns of anaphora as a case study for diagnosing sensitivity to the structure-sensitive aspects of the discourse state-updating process. 
Thus, our work provides another way of studying LLMs' state-tracking ability, through attention to the linguistic details of the discourse as opposed to the world model consequences of the actions described in a discourse. 

\subsection{Constructions} \label{subsec:cons}
We consider three operators whose scope plays a significant role in licensing discourse anaphora: universal quantifiers, negation, and disjunction.

\subsubsection{Universal Quantifiers} \label{subsubsec:uni}
\paragraph{\textit{Every}} The first case of anaphora (in)accessiblity that we consider is the universal quantifier. 
We start with a simple example, which contrasts the behavior of sentences whose subjects involve the quantifiers \textit{a} and \textit{every}. 
\pex[aboveexskip=3pt, belowexskip=3pt, interpartskip=1pt]<ae>
\a \textsc{Existential}: A farmer worked in the field.
\a \textsc{Every}: \#Every farmer worked in the field.
\a \textsc{Continuation}: He dreamed of the harvest.
\xe
As  shown in Figure \ref{fig:scopefig}, (\getref{ae}c) is felicitous following (\getref{ae}a), but not following  (\getref{ae}b). This is because the semantic scope of the existential quantifier extends indefinitely to the right, but the pronoun \textit{he} in (\getref{ae}c) is outside the scope of the universal quantifier in (\getref{ae}b).\footnote{Infelicitous examples are usually marked as \# by linguistics conventions. However, we use \# to indicate the infelicity of a sentence specifically in the context of the provided continuation.} 
In sum, the scope of universal quantifiers serves as a boundary for anaphoric accessibility. An LLM capable of discourse level understanding should therefore accurately represent the effects on the discourse context of examples like (\getref{ae}b) and reject the infelicitous continuation (\getref{ae}c).

\paragraph{\textit{Donkey Conditionals}}
\begin{figure}
    \centering
    \includegraphics[width=1\linewidth]{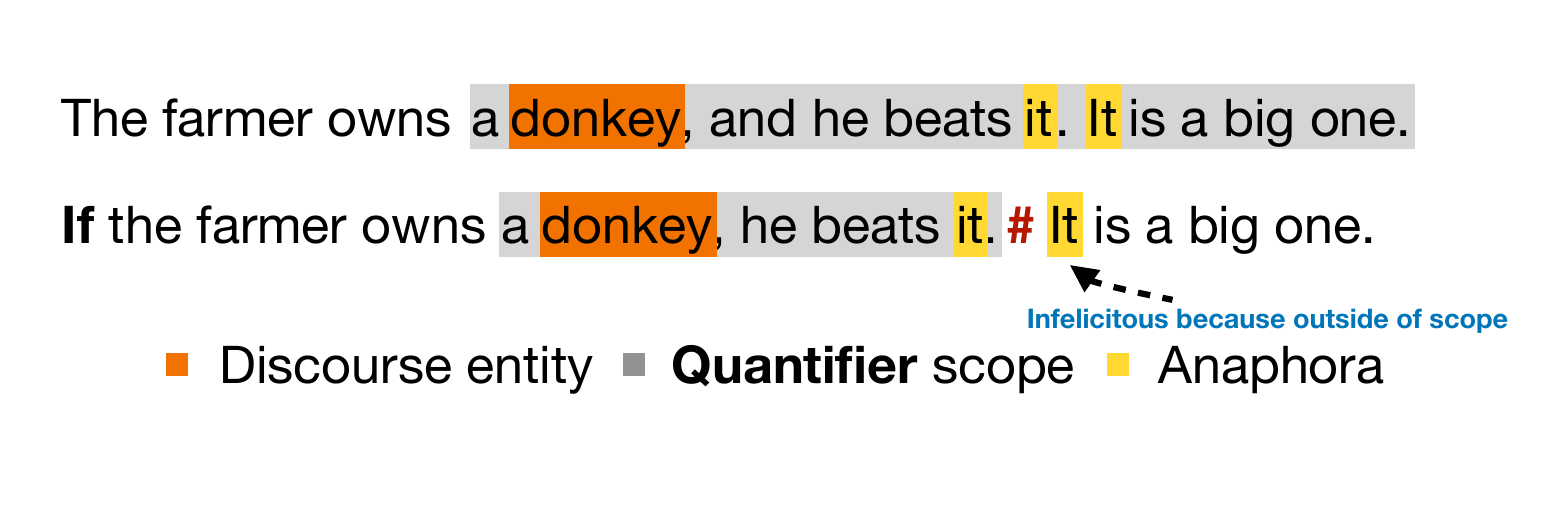}
    \caption{Illustration of anaphora accessibility in donkey conditionals.} 
    \label{fig:illus1b}
\end{figure}
A more complex case of anaphora accessibility  is known as `donkey conditionals' in the dynamic semantics literature \citep{kanazawa1994weak}. In such cases, a discourse entity is introduced via an existential quantifier in the antecedent of a conditional. In such cases, the indefinite licenses pronouns in the conditional's consequent, but not in subsequent sentences. We consider 3 cases: two types of conditional sentences, namely \textit{if} and \textit{whenever} conditionals, and conjoined sentences with an existential object in the first conjunct. 
\pex[aboveexskip=3pt, belowexskip=3pt, interpartskip=1pt]<donkey>
\a \textsc{Existential} (\textit{Exi}): John owns a donkey, and he beats it. 
\a \textsc{Conditional} (\textit{Cond}): \#If John owns a donkey, he beats it. 
\a \textsc{Whenever} (\textit{When}): \#Whenever John owns a donkey, he beats it.
\a \textsc{Continuation} (\textit{Cont}): It is a big one.
\xe
Such cases can be assimilated to the quantifier cases discussed above, if we assume the conditional clauses implicity introduce a universal quantifier that is not directly tied to a lexical quantifier (see Figure \ref{fig:illus1b}). 
Assuming this to be the case, the pronoun \textit{it} in (\getref{donkey}d) is outside the scope of the implicit universal quantifier in (\getref{donkey}b) and (\getref{donkey}c), rendering the continuation (\getref{donkey}d) infelicitous. The same continuation, however, is acceptable in (\getref{donkey}a) for the same reasons as in why (\getref{ae}a). Thus, determining that this continuation sentence is infelicitous after (\getref{donkey}b) and (\getref{donkey}c) requires accurate processing of the context sentence in preparation for the continuation and subsequent integration, which is exactly what we define as understanding at the discourse level.

\subsubsection{Negation} \label{subsubsec:negation}
Negation is another logical connective that modulates anaphora accessibility---in general, it is impossible to refer back to discourse referents that are introduced within its scope. However, double negation is an exception (see \citealt{hofmann2024anaphoric} for discussion and references). 
\pex[aboveexskip=3pt, belowexskip=3pt, interpartskip=1pt]<dne>
\a \textsc{Existential} (\textit{Exi}): The farmer owned a cow. 
\a \textsc{Negation} (\textit{Neg}): \#The farmer didn't own a cow. 
\a \textsc{DoubleNegation} (\textit{DN}): It was not the case that the farmer didn't own a cow.
\a \textsc{Continuation} (\textit{Cont}): (In fact,) It was (just) away on the meadow.
\xe
Consider the four conditions (\getref{dne}a-c) with negation, each followed by the same continuation (\getref{dne}d):
\begin{figure}
    \centering
    \includegraphics[width=1\linewidth]{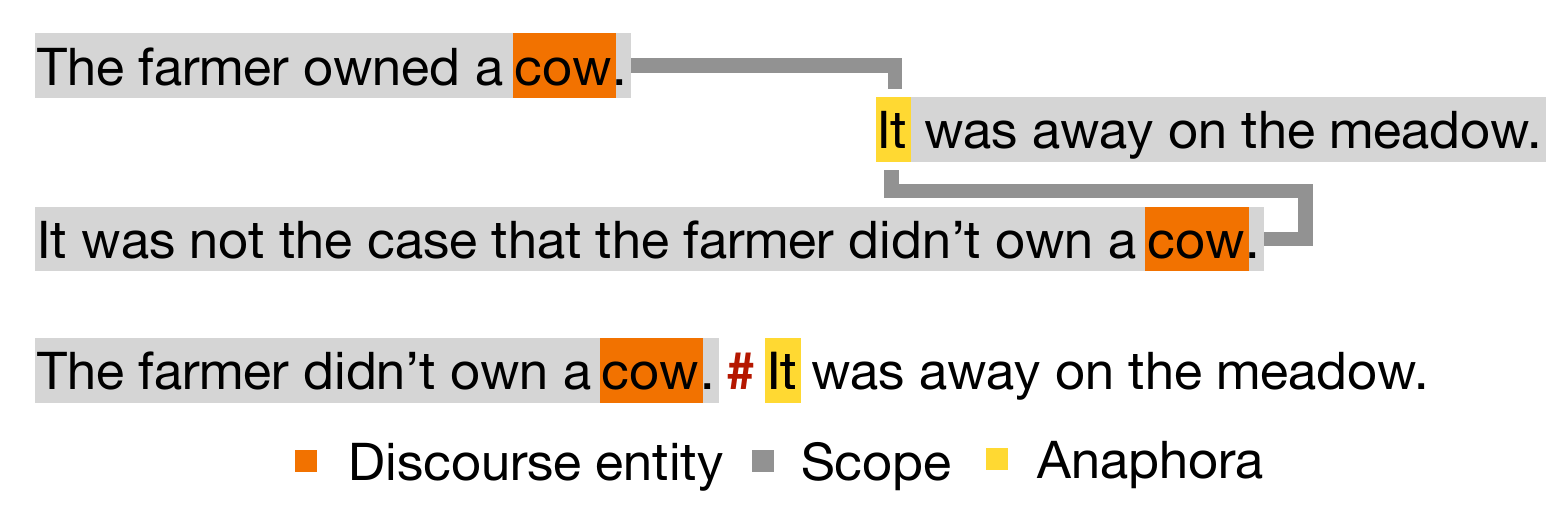}
    \caption{Illustration of anaphora accessibility in negation cases.}
    \label{fig:illus2}
\end{figure}
As is analyzed by \citeauthor{hofmann2024anaphoric} and illustrated in Figure \ref{fig:illus2}, the local context of the \textit{cow} referent in \textsc{DoubleNegation} is veridical, and the speaker is committed to the existence of \textit{a cow} owned by \textit{the farmer}. In other words, two negations cancel each other out. Thus, \textsc{Existential} is semantically equivalent to \textsc{DoubleNegation}, and both of them license the anaphora \textit{it} in \textsc{Continuation}. In contrast, no discourse referent of \textit{a cow} exists outside the scope of negation in \textsc{Negation}, which makes it an infelicitous context for the subsequent anaphora.
Here, we examine whether LLMs know the semantic scope of negation and whether negation's inaccessibility can be reversed in double negation contexts.


\subsubsection{Disjunction} \label{subsubsec:disjunction}
Negation within disjunctions adds another layer of complexity to anaphora accessibility. \citet{Evans:1977} observes that discourse referents introduced  through existentials within a first disjunct do not license anaphora in the second disjunct. Surprisingly, however, a discourse referent introduced with a negative quantifier in a first disjunct does. We see this contrast in the first two examples of (5):


\pex[aboveexskip=3pt, belowexskip=3pt, interpartskip=1pt]<andor>
\a \textsc{EitherPosOr}: \#Either there was a manuscript, or it was hidden by the librarian.
\a \textsc{EitherOr}: Either there was no manuscript, or it was hidden by the librarian.
\a \textsc{Or}: There was no manuscript, or it was hidden by the librarian.
\a \textsc{Conjunction}: \#There was no manuscript, and it was hidden by the librarian.
\xe 
\begin{figure}
    \centering
    \includegraphics[width=1\linewidth]{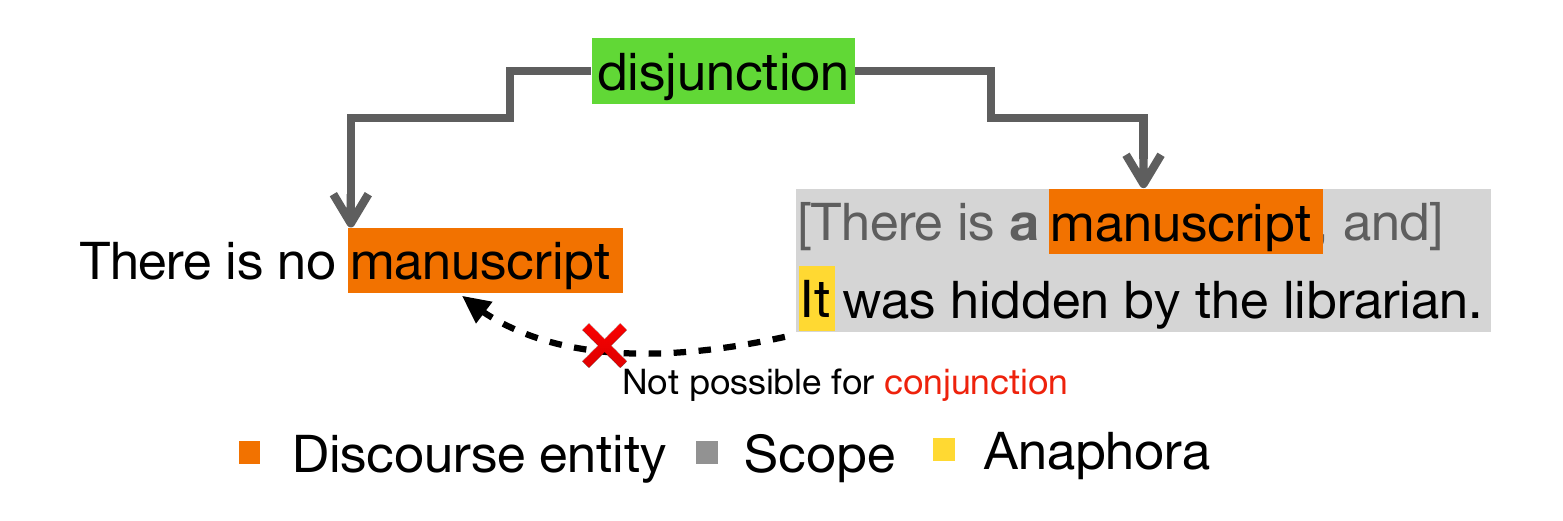}
    \caption{Illustration of anaphora accessibility in disjunction cases.}
    \label{fig:illus3}
\end{figure}
(\getref{andor}c) demonstrates that the presence or absence of the lexical item \textit{either} to introduce the disjunct does not have any impact on the discourse semantics. Finally, (\getref{andor}d) shows that negative quantifiers in conjunction do not have similar effects. 

\subsection{Experiment Design}

\paragraph{Model} We investigated the performance of four open-source LLMs (\texttt{Llama3-2-1B}, \texttt{Llama3-2-3B}, \texttt{Llama3-1-8B} and \texttt{Llama3-1-8B-Instruct} \citep{dubey2024llama}), and two closed-source LLMs (\href{https://platform.openai.com/docs/models/gpt-base}{GPT \texttt{babbage-002} and \texttt{davinci-002}}) on our constructed dataset through the Huggingface transformer API \citep{wolf2019huggingface} and the OpenAI API respectively.\footnote{We were also not able to examine more recent OpenAI LLMs such as GPT-4o because the API for these models does not support access to the log probabilities. However, the perspective and evaluation tasks we propose in this paper are still helpful in informing the discourse-level semantic understanding of state-of-the-art LLMs.} We ran inference using an NVIDIA A100 GPU with 32GB of memory allocated.

\paragraph{Human Experiment}
To establish a human baseline for models' performance, we recruited 104 participants over Prolific. Each participant did 66 forced-choice trials, with 22 experimental items and 44 fillers. In each trial, participants were visually presented with 2 minimally different sentences on the screen, and they were asked to choose the more acceptable sentence from the pair. 
See Appendix \ref{app:human} for more details on our experiment design. Human results are presented in the following sections along with language model performance.

\paragraph{Corpus}

Experimental stimuli were generated from a set of structural templates containing the target constructions. For each experiment, we manually constructed 32 semantically plausible simple sentence frames with the help of GPT-4o \citep{openai2024gpt4ocard}, following the example sentences shown in Section~\ref{subsec:cons}. Test sentences were then manually inspected by linguistics experts to ensure semantic plausibility and (un)acceptability. This yields a set of 9816 experimental sentences in total.

\paragraph{Metrics} \label{subsec:metrics}
We adopt the evaluation paradigm in \citet{futrell-etal-2019-neural} that considers LLMs as psycholinguistic subjects. That is, for each evaluated sentence, we take the surprisal (i.e., the negative log probability) assigned by the model to individual tokens, defined in Equation~\ref{eq:surprisal}:
\begin{equation}\label{eq:surprisal}
    surprisal(w_i) = \log \frac{1}{P(w_i | w_1, ..., w_{i-1})}
\end{equation}
The total probability the model assigns to a sentence or part of a sentence is obtained by taking the sum of $surprisal(w_i)$ for each target token $w_i$. The surprisal values serve as the base measurement for the analyses of each individual experiment described in the following sections. 

\begin{figure*}[t]
    \centering
    \includegraphics[width=1\linewidth]{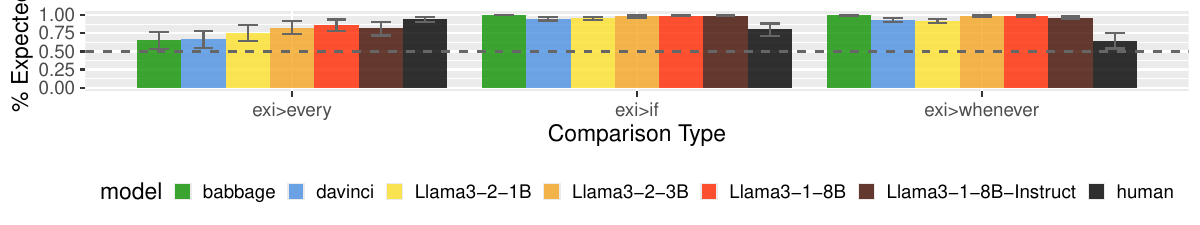}
    \caption{LLMs' performance on the comparisons involving existential vs. universal quantifiers. In the figures of this paper, $>$ signs indicate degrees of felicity. For example, \texttt{exi>every}, the label for the leftmost panel, means that \textsc{existential} should be more felicitous than \textsc{every} sentences in the relevant comparison. Such felicity preference is determined by whether models exhibit the inequality shown in equation (5).}
    \label{fig:exp1}
\end{figure*}

\section{Experiment 1: Universal} \label{sec:exp1}

In this section, we discuss models' performance on anaphora accessibility with regard to the universal quantifier as discussed in Section \ref{subsubsec:uni}. 

In general, given different context sentences and the same continuation, we expect models to assign a higher conditional probability to the continuation given a context in which it is felicitous than another context in which it is infelicitous. In other words, we expect the following inequalities to hold if LLMs exhibit discourse level understanding abilities with regard to universal quantifiers.
\begin{gather}
    p(\text{\textit{Cont}}|\text{\textit{Exi}}) > p(\text{\textit{Cont}}|\text{\textit{Every}})\\
    p(\text{\textit{Cont}}|\text{\textit{Exi}}) > p(\text{\textit{Cont}}|\text{\textit{Cond}})\\
    p(\text{\textit{Cont}}|\text{\textit{Exi}}) > p(\text{\textit{Cont}}|\text{\textit{When}})
\end{gather}
However, one problem about this measure is that it is too lenient -- although continuations such as (\getref{ae}c) are infelicitous after (\getref{ae}b), it should become felicitous if \textit{he} is instead embedded inside the scope of (\getref{ae}b), such as the contrast below.
\pex[aboveexskip=3pt,belowexskip=3pt]<inscope>
\a \textsc{CrossSen}: Every farmer worked in the field. \#He dreamed of the vest.
\a \textsc{SingleSen}: Every farmer worked in the field before he dreamed of the harvest.
\xe
Therefore, we would expect models to assign a higher probability to (\getref{inscope}b) than (\getref{inscope}a). Importantly, the contrast in example (\getref{inscope}) does not exist for their counterparts with the existential quantifier---we would expect a smaller difference in probability between them if the LLMs that we tested have good discourse level understanding abilities. Thus, instead of using equations (2), (3), and (4) as our metric, we adopt the difference-of-difference metric with the general form shown in (5). We binarize the comparison of each trial by recording whether the inequality holds in the predicted direction.

\begin{equation}\begin{split}\label{eq:ae}
    p(\exists \text{-} \textsc{SingleSen}) &- p(\exists \text{-} \textsc{CrossSen}) \\
    & < \\
    p(\forall \text{-} \textsc{SingleSen}) &- p(\forall \text{-} \textsc{CrossSen})
    \end{split}
\end{equation}

\paragraph{Results} As is shown in Figure \ref{fig:exp1}, all models show above chance performance for the expected inequality in equation (5). Specifically, for the simple comparison between \textsc{existential} and {\textsc{every}} (leftmost panel in Figure \ref{fig:exp1}), we found that the Llama family models that we tested achieved higher accuracy (around 75\%) than \texttt{babbage} and \texttt{davinci} in the GPT family, while humans scored even higher at ceiling. In the other two comparisons where the universal quantifier is implicitly encoded through \textsc{conditional} and \textsc{whenever}, LLMs continue to score at ceiling. In contrast, humans had lower accuracy but still performed above chance.
This pattern indicates that the LLMs examined know the scope of the discourse entity introduced within the universal quantifier and that it is infelicitous to refer back to such entities outside of the scope.

\begin{figure}[h]
    \centering
    \includegraphics[width=1\linewidth]{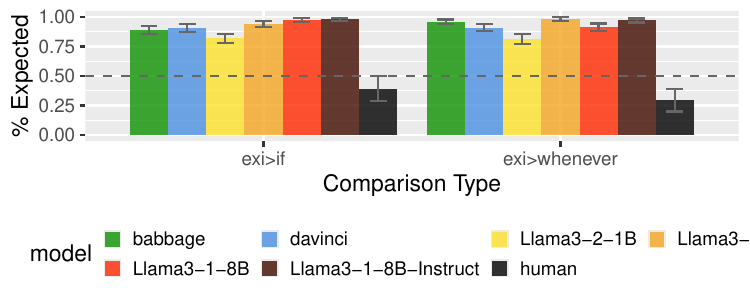}
    \caption{Model performance on \textit{he}-continuations for \texttt{exi>if} and \texttt{exi>whenever}.}
    \label{fig:exp1he}
\end{figure}

In addition to the continuation in (\getref{donkey}d) that starts with \textit{it}, for the comparisons \texttt{exi>if} and \texttt{exi>whenever}, we also considered a variant where the continuation starts with \textit{he}, such as \textit{He also feeds it}. Given our framing of the anaphora accessibility task, there should not be a difference between \textit{he}-continuations and \textit{it}-continuations---they should both be infelicitous given a preceding \textsc{conditional} or \textsc{whenever} context. Results on this variant are shown in Figure \ref{fig:exp1he}. Interestingly, there is a striking contrast between human and models' performance. While models continue to exhibit the preference for \textsc{existential} over \textsc{conditional} and \textsc{whenever}, humans actually prefer the universal counterparts for donkey conditionals, which is not predicted in the literature. We believe that this discrepancy could be due to an effect called \textit{telescoping} \citep{roberts1989modal}. The intuition is that humans have the tendency to interpret \textit{he}-continuations as being subordinated under the scope of \textsc{conditional} or \textsc{whenever}, which makes \textit{he}-continuations more felicitous than they should be. In comparison, \textit{it}-continuations are less likely to be interpreted in a subordinated way. Another potential factor that might contribute to the human performance difference between \textit{he}- and \textit{it}-continuations is subject bias: since \textit{the farmer} is the subject of the context sentence, it is more saliently represented in the discourse. Therefore, humans are more likely to refer back to it in the continuation using \textit{he}. 
In sum, the models' success on this dataset shows their knowledge of the difference between universal and existential quantifiers.

\section{Experiment 2: Negation}  \label{sec:exp2}
As discussed in Section~\ref{subsubsec:negation}, the second construction that we are interested in is negation.
Following the reasoning there, we expect the following two inequalities to hold if the LLMs understand the semantic scope of negation:
\begin{gather}
    p(\text{\textit{Cont}}|\text{\textit{Exi}}) > p(\text{\textit{Cont}}|\text{\textit{Neg}})\\
    p(\text{\textit{Cont}}|\text{\textit{DN}}) > p(\text{\textit{Cont}}|\text{\textit{Neg}})
\end{gather}
Since every pair of sentences we compare shares the continuation but not the context sentences, we apply the conditional probabilities metric: compare the summed surprisal on tokens in the \textsc{Continuation}, with the concatenated context fed to the model as a preamble.

\begin{figure}
    \centering
    \includegraphics[width=1\linewidth]{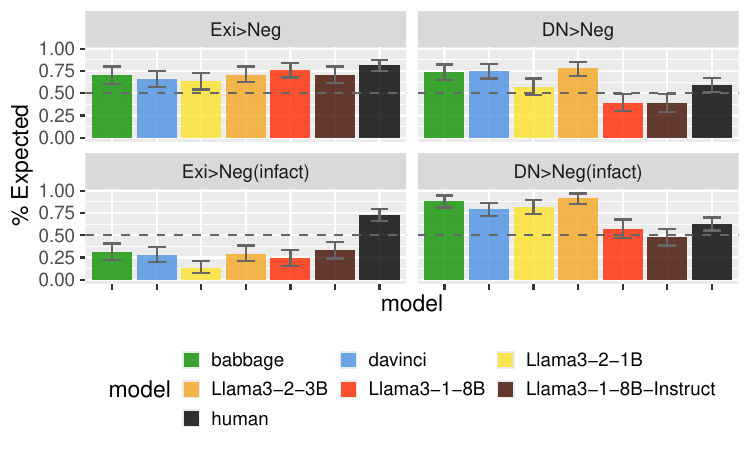}
    \caption{Model performance in Experiment 2.}
    \label{fig:exp2}
\end{figure}

\paragraph{Results}
As shown in the top two panels of Figure~\ref{fig:exp2}, all models succeed in preferring the \textsc{Existential} context over \textsc{Negation}, but three of the models struggle to favor \textsc{DoubleNegation} over \textsc{Negation}. In particular, the two \texttt{Llama3-1-8B} models show a preference of \textsc{Negation} over \textsc{DoubleNegation}, which is the reverse of what is expected. Human results, on the other hand, are high in \texttt{Exi>Neg} and exhibit a similar decrease from \texttt{Exi>Neg} to \texttt{DN>Neg}, but both are reliably above chance.
The most straightforward way to interpret these results is that the LLMs have trouble understanding that \textsc{Existential} is equivalent to \textsc{DoubleNegation} in terms of their power in licensing subsequent anaphora to discourse referents introduced within their scopes. However, another hypothesis is that \textsc{DoubleNegation} is dispreferred not because the LLMs failed to learn double negation elimination, but simply because \textsc{DoubleNegation} sentences have a more complex (and presumably less frequent) structure than its \textsc{Existential} counterpart.

\paragraph{Influence of Specific Lexical Items}
To test this hypothesis, we considered a variant of the test sentences by adding the phrase \textit{in fact} to the beginning of each continuation sentence and computed accuracy using the same inequalities as in (6) and (7). 
The intuition is that adding this phrase helps the models to better process \textsc{DoubleNegation} sentences to a larger degree than to process \textsc{Existential} ones.
If the low accuracy that we observed for the \texttt{DN>Neg} comparison is due to lexical-level factors, we would expect an increase in accuracy in the variants. In contrast, if models failed to learn the difference between double negation and negation completely, the accuracy of the variants would remain low.

\begin{figure*}[ht]
    \centering
    \includegraphics[width=1\linewidth]{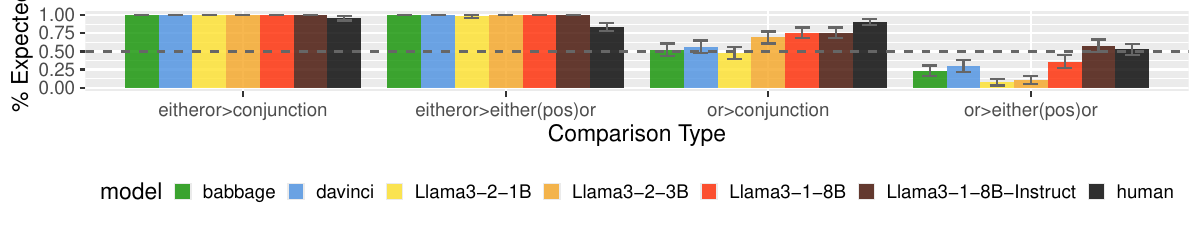}
    \caption{Model performance in Experiment 3.}
    \label{fig:exp3}
\end{figure*}

Results are shown in the bottom two panels of Figure~\ref{fig:exp2}. Compared to the base case, adding \textit{in fact} does help to lift the accuracy for the \texttt{DN>Neg} comparison, as most models now have a stronger preference of \textsc{DoubleNegation} over \textsc{Negation}. However, adding \textit{in fact} also flips the direction of the \texttt{Exi>Neg} comparison, as all models now favor \textsc{Negation} over \textsc{Existential} sentences. In contrast, human patterns remain stable regardless of the addition of \textit{in fact}: they still show a clear preference for \textsc{Existential} and \textsc{DoubleNegation} over \textsc{Negation}.

One way to interpret the flipped result is that the phrase \textit{in fact} tends to co-occur with double negation sentences, thereby increasing the conditional probabilities of the continuation. Adding \textit{in fact} to existential sentences makes the discourse less coherent to process, thereby lowering the accuracy in the \texttt{Exi>Neg(infact)} comparison. This results in the reversed \textsc{DoubleNegation}$>$\textsc{Negation}$>$\textsc{Existential} ranking by language models.
Although adding \textit{in fact} to the continuation does not change anaphora accessibility, the increase that we observed here suggests that LLMs are sensitive to the presence of specific lexical items and that their performance with respect to identifying the scope of negation is not systematic.

\section{Experiment 3: Disjunction}  \label{sec:exp3}
In the last experiment, we test the constructions presented in Section \ref{subsubsec:disjunction} with respect to disjunction. Since the sentences that we compare share neither the context nor the continuation, we calculate the Syntactic Log-Odds Ratio score (\texttt{SLOR}) \citep{lau2017grammaticality} on each sentence and compare the \texttt{SLOR} scores, which is defined as:
\begin{equation}\label{eq:slor}
    \texttt{SLOR}(s) = \frac{\log p_m(s) - \sum_{w\in s} \log p_u(w)}{|s|}
\end{equation}
where for sentence $s$, $\log p_m(s)$ represents the log probability assigned by the model to the entire sentence (which is equivalent to summing up the surprisals for all tokens in $s$); $\log p_u(w)$ represents the unigram probability of each token $w$ in the sentence; and $|s|$ represents the length of the sentence, which is the number of tokens in $s$. Intuitively, the \texttt{SLOR} score measures how much \textit{additional} probability the model assigns to the sentence compared to the same bag-of-word, which in turn represents the well-formedness of the sentence, both syntactically and semantically. However, there is no standard on how to interpret the absolute values of the \texttt{SLOR} scores. In the current study, we obtain the estimation of the unigram probabilities by counting the frequency of the tokens from a fragment of the OpenWebText Corpus \citep{Gokaslan2019OpenWeb} obtained from the tokenizers of the Llama3 family and the GPT3 family, respectively.\footnote{See the GitHub link for the unigram probability results.}

Recall from Section \ref{subsubsec:disjunction} that \textsc{or} and \textsc{EitherOr} are felicitous, while \textsc{Conjunction} and \textsc{EitherPosOr} are not. Translating the judgments to the metric, we expect the following four inequalities to hold if models exhibit discourse level understanding abilities.
\begin{gather}
    \texttt{SLOR}(\textsc{or}) > \texttt{SLOR}(\textsc{Conjunction})\\
    \texttt{SLOR}(\textsc{EitherOr}) > \texttt{SLOR}(\textsc{Conjunction})\\
    \texttt{SLOR}(\textsc{or}) > \texttt{SLOR}(\textsc{EitherPosOr})\\
    \texttt{SLOR}(\textsc{EitherOr}) > \texttt{SLOR}(\textsc{EitherPosOr})
\end{gather}

\paragraph{Results} As shown in Figure \ref{fig:exp3}, models achieved ceiling performance for all comparisons involving \textsc{EitherOr}---they demonstrate a preference for this felicitous case over \textsc{conjunction} and \textsc{EitherPosOr}, which is consistent with human preferences. In contrast, the performance is around chance for the \texttt{or>conjunction} comparison, while humans show the predicted preference pattern to a larger extent than all LMs. Strikingly, models exhibit a preference for \textsc{EitherPosOr} over \textsc{or} (rightmost panel), which is the reverse pattern of what we expect. Humans show no clear preference in this comparison. Overall, the pattern here repeats Experiment 2 in that LLMs' ability to differentiate contexts with different anaphora accessibility depends largely on lexical items and is not systematic---although \textsc{EitherOr} and \textsc{or} are equivalent to each other, models' preference largely depends on whether there is \textit{either} in the sentence.

\section{Conclusion}
In this paper, we defined a hierarchy of semantic understanding abilities consisting of lexical, sentence, and discourse levels. Filling in the gap in the literature, we constructed an evaluation task of anaphora accessibility that allows for a fine-grained examination of the understanding abilities of LLMs. Results show that our task successfully identified places of convergence and divergence between model and human performance, where LLMs rely on specific lexical cues but humans don't. This work is one further step toward improving the discourse understanding abilities of LLMs.

\section*{Limitations} 
\paragraph{Running the Dataset in SOTA Models} In the current study, we only tested our datasets with a limited range of LLMs. It would be interesting to see the performances of state-of-the-art language models such as GPT-4o and the DeepSeek model family. The main reason impeding us from testing our dataset on the latest models is that we require access to the logits the models assign to each token, which are not available for the closed-source models. Future studies could consider an alternative version of conducting such evaluation with prompting-based methods.

\paragraph{Evaluating More Subtle Constructions from Theoretical Predictions}
In addition to the three classes of quantifiers and logical connectives, there is a rich pool of linguistic constructions from the theoretical semantics literature that involve more complex scopal interactions that lead to other predictions about anaphora accessibility. An example is modal subordination (e.g., \citealp{roberts1989modal}, where the scope of \textit{if}-conditional sentence interacts with modal operators. There are few empirical studies on how humans process such sentences. Future work could further extend our dataset to incorporate a larger variety of constructions and acquire a human baseline.

\paragraph{Behavioral versus Mechanistic Level Evaluations}
In Section~\ref{sec:hierarchy}, we reviewed related works (\citealp{kim-schuster-2023-entity, li-etal-2021-implicit}) that explicitly investigate the state or discourse entity tracking capability by probing the internal activation states of language models. The current study, despite investigating the discourse updates within natural language instead of simulating discourse updates, remains at the behavioral level and is empirical in nature. Developing methods that explicitly target models' internal representations that correlate with state-update behaviors would bring greater interpretability and could contribute to theory building. Future work could improve our understanding of the processing level details of models on the current dataset by importing techniques from mechanistic interpretability.

\newpage
\bibliography{anthology,custom}

\appendix


\section{Human Experiment} \label{app:human}
\begin{figure*}
    \centering
    \includegraphics[width=1\linewidth]{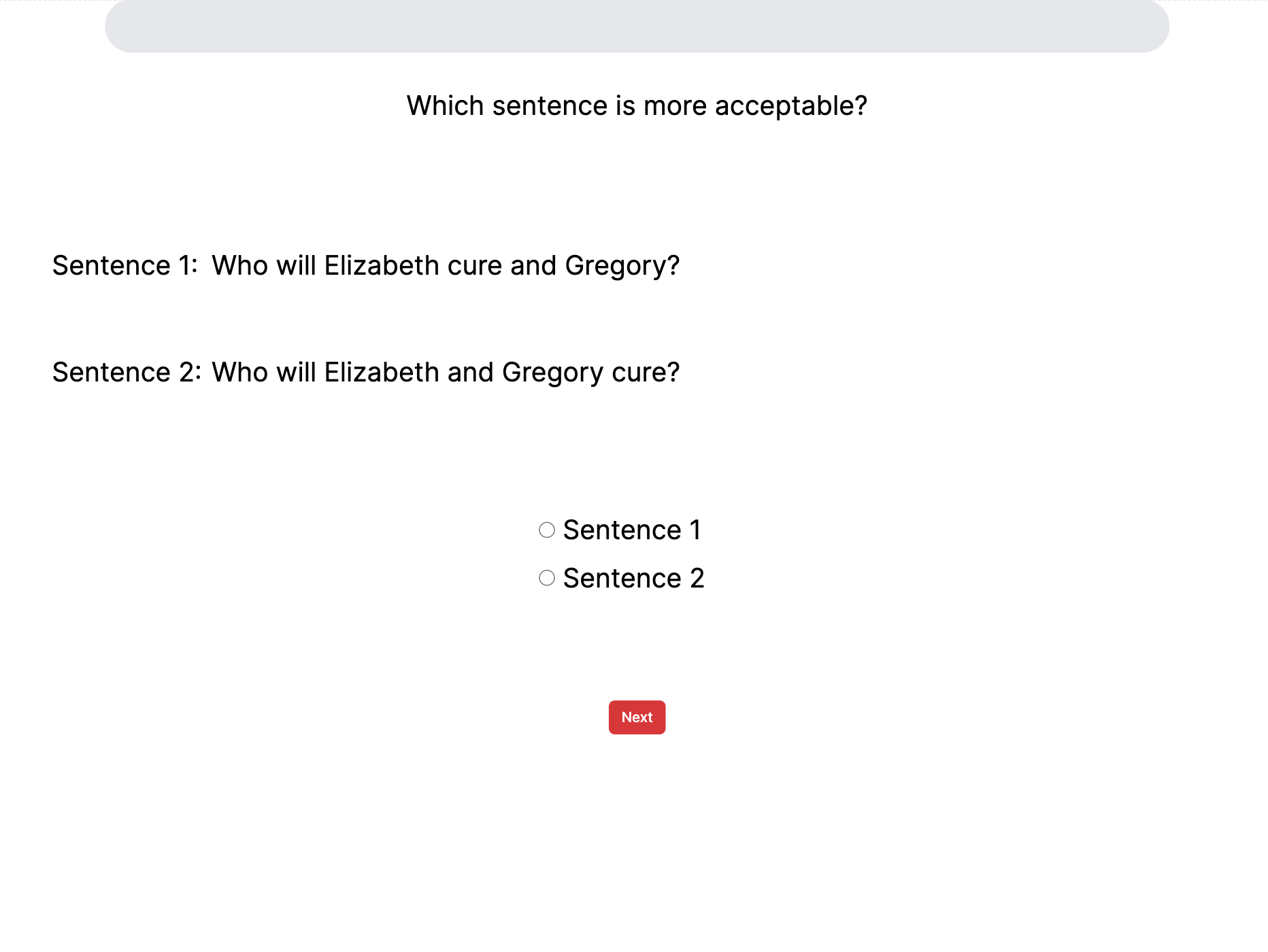}
    \caption{Experimental interface on Gorilla with an example filler item where participants were expected to click on Sentence 2.}
    \label{fig:interface}
\end{figure*}

We tested a total of 11 comparison types (3 in Experiment 1, 4 each in Experiments 2 and 3) on human subjects. Each comparison type includes 32 sentence pairs. In each test trial, participants were presented with a pair of sentences in a multiple choice format (see Figure \ref{fig:interface} for the experimental interface) and were asked to click on the sentence that they found to be more acceptable. Each participant received 22 test items and 44 filler items, which sums to a total of 66 trials. The filler items were the same across participants and were selected from BLiMP \citep{warstadt-etal-2020-blimp-benchmark} such that for each filler minimal pair, one of the sentences is strictly more acceptable than the other. Therefore, we also used filler items as attention checks. Participants who scored below 90\% accuracy on the filler items were excluded from the final results. The experiment was also set up such that each test item was rated by at least 5 participants.

We used the Gorilla Experiment Builder (\url{www.gorilla.sc}) to create and host our experiment interface \citep{anwyl-irvine_gorilla_2020}, and participants were recruited through Prolific (\url{www.prolific.com}) under a university-approved IRB. We recruited a total of 104 native speakers of English without any language or vision-related disorders who also currently reside in the United States. 85 of them (81.73\%) passed the filler check. Each participant filled out a consent form prior to completing the experiment. They each received a compensation of $\$3$, which is equal to an hourly rate of \$14.41.

\end{document}